\documentclass{article}
\usepackage{spconf,amsmath,graphicx,bm}
\usepackage{multirow}
\usepackage{graphicx}   
\usepackage{makecell}
\usepackage{subfigure}
\usepackage{cite}
\usepackage{amssymb}
\usepackage{setspace}
\pdfoutput=1

\title{3D-CSL: self-supervised 3D context similarity learning for Near-Duplicate Video Retrieval}
\name{Rui Deng$^*$, Qian Wu$^*$, Yuke Li}
\address{NetEase Yidun AI Lab, Hangzhou, China}
\email{\{dengrui01, wuqian05, liyuke\}@corp.netease.com}
\begin{document}
\ninept
\maketitle

\newcommand\blfootnote[1]{%
\begingroup
\renewcommand\thefootnote{}\footnote{#1}%
\addtocounter{footnote}{-1}%
\endgroup
}

\blfootnote{$^*$These authors contributed equally to this work.}

\vspace{-2em}
\begin{abstract}
In this paper, we introduce \textbf{3D-CSL}, a compact pipeline for Near-Duplicate Video Retrieval (NDVR), and explore a novel self-supervised learning strategy for video similarity learning. Most previous methods only extract video spatial features from frames separately and then design kinds of complex mechanisms to learn the temporal correlations among frame features. However, parts of spatiotemporal dependencies have already been lost. To address this, our 3D-CSL extracts global spatiotemporal dependencies in videos end-to-end with a 3D transformer and find a good balance between efficiency and effectiveness by matching on clip-level. Furthermore, we propose a two-stage self-supervised similarity learning strategy to optimize the entire network. Firstly, we propose \textbf{PredMAE} to pretrain the 3D transformer with video prediction task; Secondly, \textbf{ShotMix}, a novel video-specific augmentation, and \textbf{FCS loss}, a novel triplet loss, are proposed further promote the similarity learning results. The experiments on FIVR-200K and CC\_WEB\_VIDEO demonstrate the superiority and reliability of our method, which achieves the state-of-the-art performance on clip-level NDVR. \textbf{Code will be publicly available.}

\end{abstract}
%
\section{Introduction}\label{sec:intro}
\vspace{-0.3em}
With the increasing proportion of video data on the Internet, video retrieval algorithms play a critical role in various scenarios, such as recommendation, search, de-duplication, etc. Previous methods pre-extract spatial features frame-by-frame with 2D networks and can be divided into frame-level/fine-grained and video-level/coarse-grained video retrieval. Frame-level methods\cite{kordopatis2019visil,shao2021tca,liu2017image,chou2015pattern} finely compare every pair of frames and compute similarities between two videos by designed mechanisms. The retrieval processes of frame-level methods are usually slow due to the fine-grained mechanism and cost a large amount of storage space because all the frame features must be stored. Video-level methods are proposed to further aggregate frame-level features so that every video can be represented by only one vector\cite{kordopatis2017dml,lee2020large}, hash code\cite{song2011multiple,song2018selfhashing,yuan2020central}, or other forms of indexes\cite{kordopatis2017near,liang2020efficient}, which are compact but may be insufficient to describe the details of the video. There is a trade-off between efficiency and effectiveness. So VRL\cite{he2022vrl} is proposed to use clip-level features to represent a fixed-length video segment. Besides, temporal context information is essential for reducing the redundant information between frames and acquiring a determinate representation. Although some of the above methods\cite{shao2021tca,he2022vrl} design modules to extract the temporal correlations between frame features, spatiotemporal dependencies are partially lost while extracting the frame features separately.

\begin{figure}[t]
  \centering
  \centerline{\includegraphics[width=8cm]{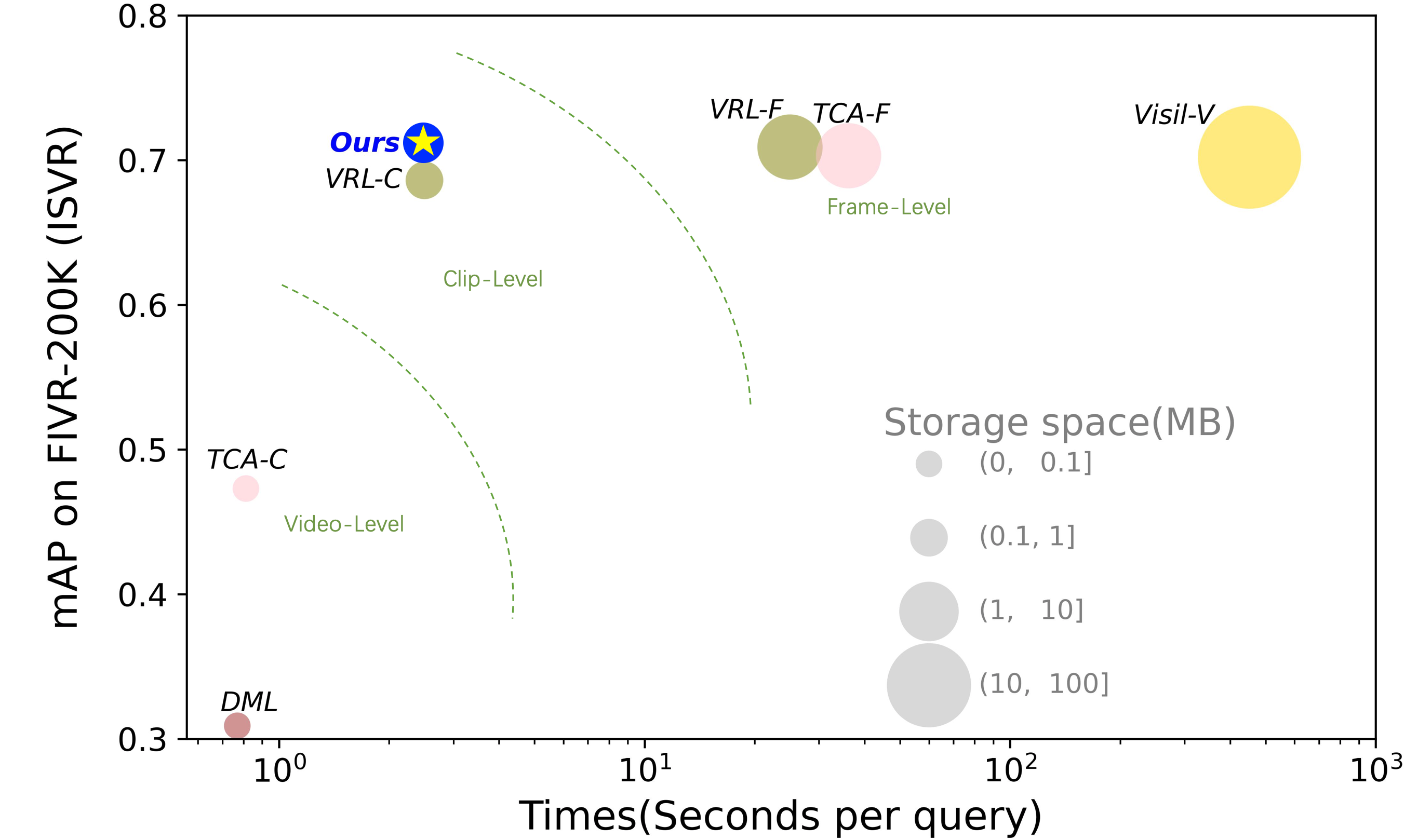}}
  \vspace{-1em}
\caption{Video Retrieval performance comparison on
FIVR-200K\cite{kordopatis2019fivr} in terms of mAP, time cost, and feature storage space. Our 3D-CSL achieves the best trade-off between efficiency and effectiveness against SOTA methods.}
\vspace{-1.5em}
\label{fig:performance}
\end{figure}

Therefore, this paper proposes a compact 3D clip-level pipeline to address the abovementioned problems. At first, we adopt the clip-level retrieval scheme as VRL\cite{he2022vrl} and split a video into several clips. And then, we extract spatiotemporal video features by a 3D transformer instead of frame-level spatial features. The final similarities are computed with these clip features, with no additional modules needed for obtaining temporal context, ensuring an extremely compact pipeline for video retrieval.

Benefiting from the compact design, we can optimize the entire pipeline with video similarity learning. Most of the previous works\cite{kordopatis2017dml,shao2021tca,kordopatis2019visil} use pair-wise methods to supervise model optimization. Some\cite{kordopatis2017dml,kordopatis2019visil} use triplet loss, which covers limited correlations and highly depends on sampling hard negative pairs. Also, pair-wise video annotations are very expensive, so their training datasets are usually small (e.g. the annotated core set of VCDB\cite{jiang2014vcdb}). So TCA\cite{shao2021tca} takes advantage of more negative samples by constructing a memory bank referring to contrastive learning\cite{mocov1,simclrv1}. VRL\cite{he2022vrl} further proposes using large-scale unlabeled data through self-supervised learning, generating deformations of the anchor as positive samples with various spatial and temporal transforms to get rid of the dependence on annotations. They all use InfoNCE loss\cite{oord2018infonce}, a common loss in contrastive learning. However, InfoNCE loss penalizes all sample pairs equally. In contrast, hard positives (positive sample pairs with small similarities) and hard negatives (negative sample pairs with large similarities) samples should be optimized much more. So Circle loss\cite{sun2020circle} re-weights each similarity to highlight the hard samples. Multi-similarity loss (MS loss)\cite{wang2019msloss} takes account of three types of similarities and sets a principle to mine the most informative sample pairs for optimization. However, the relative magnitudes between different positive sample similarities are still ignored. In some scenarios, we want to distinguish the positive samples with different similarities according to their distances from the query. Therefore, we design the Flipped Clip Similarity Loss (FCS loss) based on the presence of Visual Chirality\cite{lin2020visualchirality}. FCS loss constrains the similarity of the non-flipped sample to be greater than the flipped one, so as to lead the model to evaluate the relative magnitudes of similarities more accurately. During similarity learning, we also propose ShotMix, a new augmentation method specific to videos, to increase random perturbations of the shot number during training, which helps the model to be more robust to extract representations for multi-shot videos.

Moreover, we employ self-supervised pretraining for better initialization. Different from many previous methods\cite{mocov1,simclrv1,o3n,cop}, VideoMAE\cite{tong2022videomae} leads the model to learn the spatiotemporal correlations by reconstructing a video clip's randomly masked pixel tubes. However, the inherent temporal properties of videos are not leveraged sufficiently, such as that the future frames can be used as the ground truth for video prediction from the past frames. Thus, we propose PredMAE to urge the model to learn a stronger temporal reasoning ability by future frame prediction task, which leads to more expressive video representations. The model initialized by pretrained weights has a notable improvement in similarity learning.

Our contributions can be summarized as follows: (1) a novel and compact 3D video retrieval pipeline. To our best knowledge, it is the first 3D transformer pipeline used for NDVR. (2) a self-supervised similarity learning strategy with \textbf{FCS loss} designed to distinguish the relative similarity magnitudes of positive samples. (3) \textbf{ShotMix}, a video-specific data augmentation, which effectively enhances the model robustness to handle the clips with various numbers of shots. (4) \textbf{PredMAE}, a self-supervised pretraining method based on future frame prediction, which effectively promotes similarity learning.

\begin{figure}[t]
    \centering
    \centerline{\includegraphics[width=8.5cm]{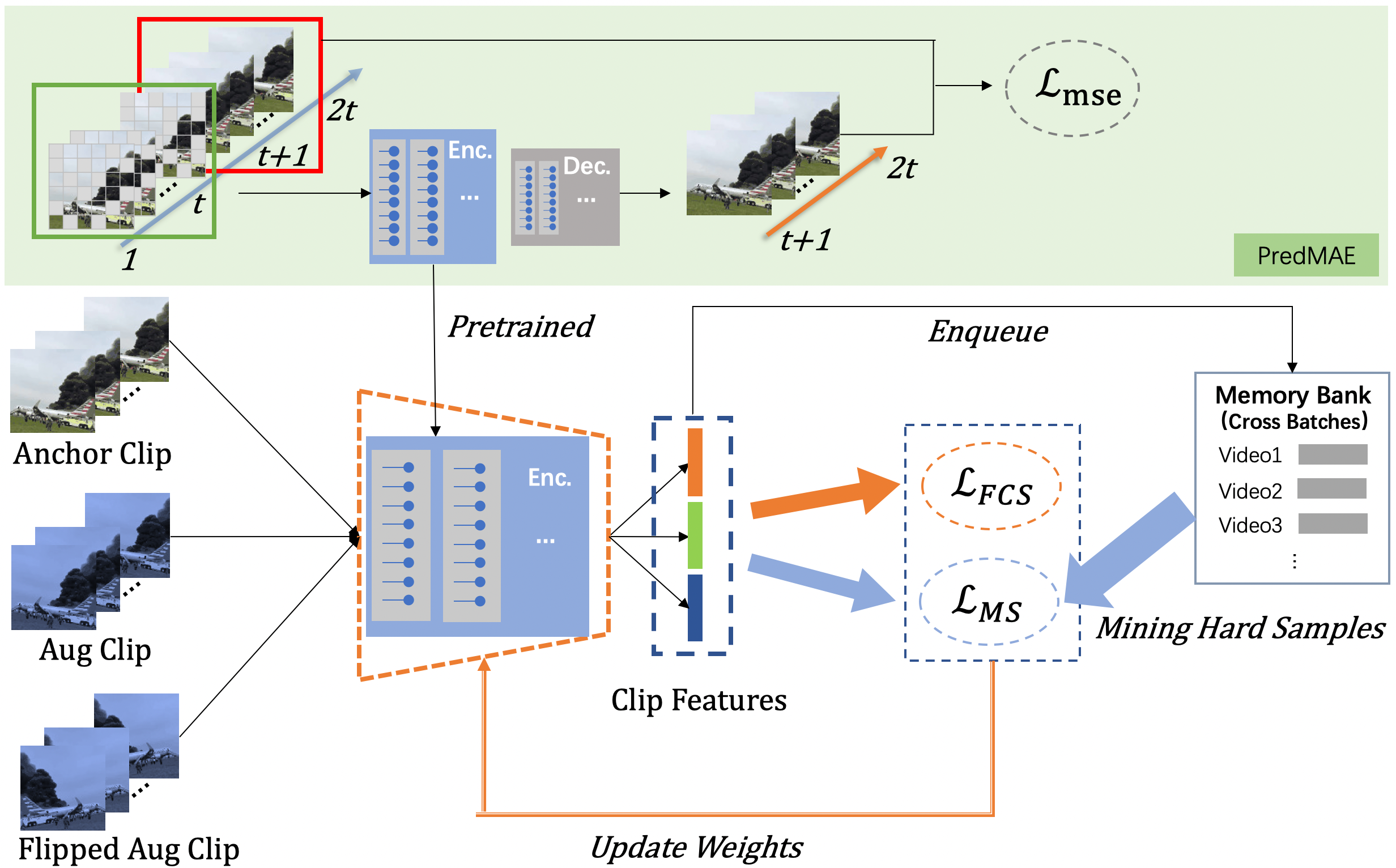}}
    \vspace{-1em}
    \caption{Overview of our Video Retrieval pipeline and training strategy. In the pretraining stage (top): the 3D encoder takes the past randomly masked $t$ frames as input and then the decoder predicts the future $t$ frames. In the similarity learning stage (bottom): the pretrained encoder extracts the features of an anchor clip and its two transformations respectively. The three clip features are used to compute the FCS loss. Hard samples are mined for computing MS loss.}
    \vspace{-1.5em}
    \label{fig:overview}
\end{figure}

\vspace{-0.5em}
\section{The Compact Video Retrieval Pipeline}
\vspace{-0.5em}
\label{sec:proposed}
\textbf{A Compact 3D pipeline.} Different from previous video retrieval works, this paper focuses on how to construct an end-to-end video retrieval pipeline, as shown in Fig.\ref{fig:overview}. We adopt a 3D backbone (Timesformer\cite{timesformer} is used in this paper) as the spatiotemporal feature extractor to avoid losing temporal information, which happens when extracting frame features separately. In order to balance the retrieval efficiency and effectiveness, we extract clip-level features\cite{he2022vrl} from video segments, which matches the scheme of the 3D video feature extractor well. Formally, the input video $V$ is split into N clips and $F$ frames are uniformly sampled from each clip. The 3D extractor takes clips $ X \in \mathbb{R}^{N \times F \times H \times W \times 3} $ as input and learns the spatial, temporal and spatiotemporal correlations among frames all at once. The output $ Y \in \mathbb{R}^{N \times D} $ is the representations with Dimension $D$ for the $N$ input clips.



\noindent\textbf{TopK Chamfer Similarity.} The similarity between two videos should be measured considering all pairs of clips, and Chamfer Similarity (CS) is generally adopted. We denote the representations of two clip sequences as $ A = [a_1, a_2, ..., a_{n}] \in \mathbb{R}^{n\times D} $ and $ B = [b_1, b_2, ..., b_{m}]\in \mathbb{R}^{m\times D}$. 
The $Sim_{CS}$ is computed as the mean of all the $n$ max similarities $\{ \max_{j\in (1,m)}(Sim_{a_i, b_j})\}_{i=1}^n $. However, some of the maximum values are actually not large enough because the pairs of clips $\{(a_i, b_j)\}_{j=1}^{j=m}$ are all dissimilar, which may lead to a smaller CS similarity than we expect. In fact, we generally focus on the most similar pair of clips \cite{gorti2022xpool}, so we develop the CS to TopK-CS, where only the top $k$ max similarities of $A$ are averaged to get the final $Sim_{topkCS}$ (we set $k=3$ in this paper):

\vspace{-0.8em}
\begin{equation}
Sim_{topkCS}=\frac{1}{k} \sum_{i=1}^{k}{sort}_{topk}(\max_{j} {a_i b_j^T})
\label{equ:top3cs}
\end{equation}
\vspace{-1em}

The 3D pipeline discards the complex temporal context learning modules in previous works\cite{kordopatis2019visil,shao2021tca,he2022vrl}, and the TopK-CS can be computed fast with only one $D$-dimension matrix multiplication, which makes the retrieval process fast.

\vspace{-0.5em}
\section{Self-supervised 3D context similarity learning strategy}
\vspace{-0.5em}
Supervised metric learning methods need positive/negative-pair annotations, which are very expensive, especially for videos. Fortunately, videos are time sequences where the future frames are natural ground truth ($GT$) for prediction from the past frames. Thus, we propose the \textbf{Predictive Masked Autoencoder (PredMAE)} to leverage this natural of videos to learn the inherent spatiotemporal correlation in video sequences with one step further based on VideoMAE\cite{tong2022videomae}, as described in Sec 3.1. Subsequently, a video-specific augmentation, \textbf{ShotMix}, and a novel triplet loss, \textbf{Flipped Clip Similarity Loss (FCS loss) }, are proposed to improve the model performance on video retrieval tasks, as described in Sec.3.2 and Sec.3.3.

\vspace{-1em}
\subsection{Predictive Masked Autoencoder (PredMAE)}
\vspace{-0.5em}
\label{sec:predmae}
By virtue of the pretext task, reconstructing the randomly masked spatiotemporal video tubes, VideoMAE\cite{tong2022videomae} successfully learned great video representations and promoted many downstream tasks. When the model $f(\cdot)$ reconstructs the masked video tubes $Q_{ts}$, the other correlated and not masked tubes in $\{Q_{11}, \cdots, Q_{FS}\}$ provide information depends on their distances from $Q_{ts}$. $F$ is the frame number, and $S$ is the spatial patch number in one frame. However, the distance $<Q_{t-1,s}, Q_{ts}>$ and $<Q_{t+1, s}, Q_{ts}>$ are the same while the direction relationship cannot be represented inadequately, which causes a lack of description about the direction of time elapsing and some motions with direction features are hard to be distinguished, such as ``standing up" and ``sitting down". Meanwhile, the long-range dependencies may be learned insufficiently because they contribute less to the tube reconstruction, but they are still vital to represent the video for downstream tasks.

Thus, we propose the PredMAE to replace the reconstruction task with future frame prediction. We randomly mask 90\% spatiotemporal tubes from clip input $X = \{x_1, x_2, \cdots, x_t\}$ following VideoMAE\cite{tong2022videomae} and get the masked clip input $X_m = \{x_{m1}, x_{m2}, \cdots, x_{mt}\}$. The next $t$ frames $X_{GT} = \{x_{t+1}, x_{t+2}, \cdots,$ $x_{2t}\}$ are the $GT$ for future frames prediction. The prediction model is composed of an encoder $\mathcal{E}(\cdot)$ and a decoder $\mathcal{D}(\cdot)$ as shown in Fig.\ref{fig:overview}. We take ViT-base/small and the divided space-time attention (namely Timesformer\cite{timesformer}) followed with an MLP as the backbone of $\mathcal{E}(\cdot)$. $\mathcal{D}(\cdot)$ is a 4-layer transformer with an MLP followed. MSE loss is adopted to constrain the prediction. Without reference from future frames, the model tends to learn the longer-range dependencies as supplement. Modeling long-range dependencies is the strength of transformer and PredMAE brings out its potential.


\vspace{-1em}
\subsection{Self-supervised 3D Context Similarity Learning}
\vspace{-0.5em}
\label{ssec:fcsloss}

In the second-stage similarity learning, we generate positive samples for a given anchor by spatial and temporal augmentations and the other samples in the batch are negative. To increase negative samples, we construct a memory bank\cite{mocov1,simclrv1} across batches and adopt the mining strategy as MS loss\cite{wang2019msloss} to obtain the informative hard samples. Considering a given anchor sample $x_a$, only hard negative samples $x_n$, with similarities greater than the minimum positive similarity $(S_{pos\_min} + margin)$, and the hard positive samples $x_p$, with similarities smaller than the maximal negative similarity $(S_{neg\_max} - margin)$, are taken into account by the MS loss: 

\vspace{-1em}
\begin{equation}
\begin{aligned}
\mathcal{L}_{MS} = \frac{1}{M}\sum_{i=1}^{M} &\{\frac{1}{\alpha}log[1 + \sum_{k\in {\mathcal{P}_i}}e^{-\alpha (S_{ik}-\lambda)}] \\
&+ \frac{1}{\beta}log[1+\sum_{k\in \mathcal{N}_i}e^{\beta(S_{ik}-\lambda})]\}
\end{aligned}
\end{equation}
\vspace{-0.9em}

\noindent where $M$ is the number of training samples, $S_{ik}$ denotes the similarity between $x_i$ and $x_k$, $\mathcal{P}_i$ and $\mathcal{N}_i$ are the mined sets of positive clips and negative clips, respectively. $\alpha$ and $\beta$ are fixed hyper-parameters. 

\noindent\textbf{Flipped Clip Similarity Loss.} 
The MS Loss only considers the similarities between the positive and negative samples, while the relative positive similarity discrimination ability of the model is also essential. For example, the DSVR task\cite{kordopatis2019fivr} only takes the edited version of the query as positive, while taking the other videos containing the same scenes as negative, although these videos can be treated as positive in other tasks. Therefore, we propose the Flipped Clip Similarity Loss (FCS loss), which drives the model to distinguish the relative positive similarities by flipping clips. The inspiration comes from Vision Chirality\cite{lin2020visualchirality}, which reveals that flipped images can bring a dramatic semantic change due to the asymmetry of vision chirality. Therefore, a flipped positive clip $x_{pf}$ is regarded to be farther away from the anchor clip $x_a$ than other positive but not flipped clips $x_{p}$. Namely, the similarity $Sim_{pf}$ between $x_{pf}$ and $x_a$ is regarded to be smaller than $Sim_p$, the similarity between $x_{p}$ and $x_a$. Thus, the FCS loss can be designed as:


\vspace{-1.7em}
\begin{equation}
\mathcal{L}_{FCS} = \frac{1}{N} \sum_i^N { \max \{ D(x_{ai}, x_{pi}) - D(x_{ai}, x_{pfi})+\gamma, 0 \}}   
\end{equation}
\vspace{-1.2em}

\noindent where $N$ is the batch size, $\gamma$ is the minimum margin between the non-flip positive pair and flip positive pair, $D(\cdot)$ means the distance between a pair of clips. Here we use Consine distance.

In this way, the three kinds of similarity discussed as in \cite{wang2019msloss} are all considered, while the self-similarities of different positive samples can be distinguished as well with the help of FCS loss. The whole loss is constructed by the two parts with weights $w_{1}$ and $w_{2}$:
\vspace{-0.5em}
\begin{equation}
\mathcal{L}= w_{1} \mathcal{L}_{MS} + w_{2} \mathcal{L}_{FCS}
\end{equation}
\vspace{-1.8em}

\noindent\textbf{ShotMix.} Different from images, the information volume in a video varies with the duration and the shot number. When a clip contains more shots, information changes more rapidly among frames while keeping nearly similar in the same shot. In similarity learning, a pair of positive clips sampled contain few shots may be easy samples since their frames are very similar. Therefore, we propose ShotMix to reconstruct such easy samples by increasing random perturbations of the shot number. Formally, given a video $V$, two frame sequences are sampled as a pair of positive clips $x_a = \{f_{ai}\}_{i=1}^F$, $x_p = \{f_{pi}\}_{i=1}^F$ with an overlapped sub-clip. The length of the overlapped sub-clip is $r * F$ and the ratio $r$ randomly ranges in $(0.7, 1.0)$. Then, we further sample a cut $x_c$ with a length ranging in  $(0, (1-r)*F)$ from $V$ to replace the beginning or the end of $x_p$ to get $x_p'$, so that more shots from $x_c$ can be mixed into it. When $x_c$ is dissimilar with $x_a$, $x_p'$ will get harder due to the mixed noise. When $x_p'$ is similar with $x_a$, it increases the difficulty for the model to describe the information in $x_p'$, which also makes $x_p'$ more useful for optimizing the video feature extractor.



\vspace{-0.5em}
\section{EXPERIMENTS}
\vspace{-0.5em}

\label{sec:majhead}
  \subsection{Experiment Setting}
  \vspace{-0.5em}
  \label{ssec:subhead}
  \textbf{Dataset.} We use Kinetics400\cite{carreira2017quo-k400} as training dataset to train the self-supervised model without using any labels. It consists of 240k videos from Youtube. For evaluation, we test the \textbf{NDVR} task on \textbf{CC\_WEB\_VIDEO}\cite{wu2007practical-ccweb} and \textbf{FIVR-200K}\cite{kordopatis2019fivr} dataset. \textbf{CC\_WEB\_VIDEO} consists of 24 query sets and 13129 videos. \textbf{FIVR-200K} dataset has 100 queries and 225960 videos and includes three different fine-grained retrieval tasks.
   
  \noindent\textbf{Training Details.} In stage 1,  we use 8 frames as input to predict the next 8 frames. The models are trained from scratch with a mask ratio of 0.9. In stage 2,  we randomly apply the base augmentation of cropping, rotation, and color-jitter. The values of $w_1$, $w_2$, and $\gamma$ in FCS Loss are 1, 0.01, and 0.1. We use the same optimizer settings on two stages with 8 A100 GPUs.  We adopt \textbf{AdamW}\cite{loshchilov2017decoupled-adamw} as optimizer and cosine annealing as learning rate scheduler. The learning rate is adjusted by  \textit{$lr=\frac{base\_lr*batch\_size}{256}$} and  $base\_lr$ is 5e-4.
  
  \noindent\textbf{Model Inference.} During the inference stage, we extract frames at a rate of one frame per second and use every 8 frames to form a clip. The dimension is 384 / 768 for Timesformer-Small / Base model.

   \begin{table}[b!]
   \vspace{-1.7em}
    \centering
    \setlength\tabcolsep{4pt}
\begin{tabular}{|l|ccc|ccc|}
\hline
\textbf{Pretrain}              & \textbf{SH} & \textbf{FL} & \textbf{TC} & \textbf{DSVR}  & \textbf{CSVR}  & \textbf{ISVR}  \\ \hline
ImageNet                       & -           & -           & -           & 0.773          & 0.729          & 0.603          \\ \hline
VideoMAE\cite{tong2022videomae}                            & -           & -           & -           & 0.796          & 0.752          & 0.624          \\ \hline
\multirow{4}{*}{PredMAE (Ours)} & -           & -           & -           & 0.810          & 0.760          & 0.635          \\
                               & Y           & -           & -           & 0.841          & 0.787          & 0.663          \\
                               & Y           & Y           & -           & 0.850          & 0.807          & 0.689          \\
                               & Y           & Y           & Y           & \textbf{0.862} & \textbf{0.818} & \textbf{0.694} \\ \hline
\end{tabular}
      \vspace{-1em}
    \caption{Effectiveness of the proposed methods on FIVR-200K. \textbf{SH}, \textbf{FL} and \textbf{TC} stand for ShotMix, FCS loss and TopK-CS respectively.} 

    \label{tab:abtest}
  \end{table}

  \begin{table*}[ht!]
  \setlength\tabcolsep{4pt}
  \begin{tabular}{c|c|ccc|ccc|cccc}
    \hline
    \multirow{2}{*}{\textbf{Retrieval}} & \multirow{2}{*}{\textbf{Method}} & \multirow{2}{*}{\textbf{Trainset}} & \multirow{2}{*}{\textbf{Use Label}} & \multirow{2}{*}{\textbf{\begin{tabular}[c]{@{}c@{}}Hours(Nums.)\end{tabular}}} & \multicolumn{3}{c|}{\textbf{FIVR-200K}}          & \multicolumn{4}{c}{\textbf{CC-WEB-VIDEO}}                         \\ \cline{6-12} 
                                       &                                  &                                   &                                     &                                                                                   & DVSR           & CSVR           & ISVR           & cc\_web        & cc\_web*       & cc\_web\_c       & cc\_web\_c*    \\ \hline
    \multirow{2}{*}{Video}             & DML\cite{kordopatis2017dml}                              & VCDB                              & Y                                   & 1999(100k)                                                                        & 0.398          & 0.378          & 0.309          & 0.971          & 0.941          & 0.979          & 0.959          \\
                                      & TCA-C\cite{shao2021tca}                            & VCDB                              & Y                                   & 1999(100k)                                                                          & 0.570          & 0.553          & 0.473          & 0.973          & 0.947          & 0.983          & 0.965          \\ \hline
   \multirow{5}{*}{Frame}             & TN\cite{tan2009scalable}                              & VCDB                                  & Y                                    & 1999(100k)                                                                     & 0.724          & 0.699          & 0.589          & -          & -         & -         & -         \\
                                    & PPT\cite{chou2015ppt}                          & -                                 & N                                   &  -           & 0.775          & 0.740          & 0.632          & 0.958          & 0.949          & -              & -              \\
                                       & TCA-F\cite{shao2021tca}                            & VCDB                            & Y                                   & 1999(100k)                                                                             & 0.877          & 0.830          & 0.703          & 0.983          & 0.969          & 0.994          & 0.990          \\
                                       & Visil-V\cite{kordopatis2019visil}                          & VCDB                              & Y                                   & 1999(100k)                                                                          & 0.892          & 0.841          & 0.702          & \textbf{0.985} & \textbf{0.971} & \textbf{0.996} & \textbf{0.993} \\
                                       
                                       & VRL-F\cite{he2022vrl}                            & YouKu\dag                            & N                                   & 3000(  -  )                                                                            & \textbf{0.900}          & \textbf{0.858}          & \textbf{0.709}          & -          & -          & -          & -          \\ \hline
    \multirow{4}{*}{Clip}              & VRL-C\cite{he2022vrl}                            & YouKu\dag                            & N                                   & 3000(  -  )                                                                            & 0.876          & \textbf{0.835} & 0.686          & -              & -              & -              & -              \\
                                       & Ours-S                           & K400                              & N                                   & 265(100k)                                                             &     0.855           &  0.812   & 0.688   & 0.970       &   0.958    &  0.982    &  0.978    \\
                                       & Ours-B                           & K400                              & N                                   & 265(100k)                   &     0.862            &     0.818            &   0.694              &      0.970          &   0.959             &    0.982            &      0.978          \\
                                       & Ours-B                           & K400                              & N                                   & 635(240k)                                                                            & \textbf{0.879} & \textbf{0.835}     & \textbf{0.711} & \textbf{0.970} & \textbf{0.959} & \textbf{0.982} & \textbf{0.979} \\ \hline
    \end{tabular}
  \vspace{-0.8em}
  \caption[]{Compare with state-of-the-art on FIVR-200K and CC\_WEB\_VIDEO. \dag The dataset is collected by author and not publicly available. }
  \vspace{-1.5em}
  \label{tab:sota}
\end{table*}

\vspace{-0.8em}
  \subsection{Ablation Study}
  \label{ssec:subheadd}
\vspace{-0.6em}
  \textbf{(1) Effectiveness of Proposed Methods.} In this section, we evaluate the retrieval performance of the two stages of 3D-CSL. We use a randomly selected subset of K400 with 100k videos as the training dataset and FIVR-200K as the evaluation dataset. The backbone is Timesformer-Base. We begin with a solid baseline model that initializes with ImageNet\cite{timesformer} and optimizes with MS-Loss\cite{wang2019msloss}. Then we illustrate the effect of using different pretrain methods. To compare with VideoMAE\cite{tong2022videomae}
  , which is similar to the proposed PredMAE, we reimplement it and replace the backbone with Timesformer. Next, we assess the effect of the proposed augmentation of ShotMix, which is used in stage 2 with a probability of 0.5. Finally, we compare the performance of with/without FCS Loss and TopK-CS.
  
As shown in Table\ref{tab:abtest}, our 3D-CSL outperforms the baseline by a large margin of \textbf{9\%}. We observe that integrating PredMAE, ShotMix, FCS Loss, and TopK-CS each contributes to improve performance. Our results suggest that the gain comes from the good pretrained features, as well as various self-supervised information generated by ShotMix and FCS Loss.

\vspace{0.2em}
\noindent\textbf{(2) Efficiency of Data and Model.} The robustness of our 3D-CSL allows it to work well on small datasets and small models. We validate this property on Timesformer-Small and a 100k subset of K400. Compared to other datasets, we notice that K400 has the shortest duration of 635 hours, which is 31\% of VCDB\cite{jiang2014vcdb} and 21\% of Youku\dag\cite{he2022vrl}. When aligned the video number with VCDB, the duration decreases to 265 hours. From Table\ref{tab:sota}, we observe that the small model is still competitive and achieves \textbf{0.855}, \textbf{0.812}, and \textbf{0.688} mAP on FIVR-200K. Besides, the 6-layer Timesformer-Small has only 50\% of parameters compared to 12-layer Timesformer-Base. 

Moreover, our models can benefit from the increased data volume and larger model size. We validate it on Timesformer-Base with the entire 240k dataset. From table\ref{tab:sota}, this model achieves an average improvement of \textbf{2.4\%} on FIVR-200K. We set this model as default model to compare against recent state-of-the-art models.




\vspace{-1em}
\subsection{Comparison to the State-of-the-Art}
\vspace{-0.5em}

We compare the performance of 3D-CSL with several state-of-the-art approaches on FIVR-200K\cite{kordopatis2019fivr} and CC\_WEB\_VIDEO\cite{wu2007practical-ccweb}. We report the best results of the original papers. For video-level methods, we report DML\cite{kordopatis2017dml} and TCA-C\cite{shao2021tca}. For frame-level methods, we report TN\cite{tan2009scalable}, PPT\cite{chou2015ppt}, TCA-F\cite{shao2021tca}, Visil\cite{kordopatis2019visil}, and VRL-F\cite{he2022vrl}. As for clip-level methods, such methods have been proposed recently, and we have found only one method (VRL-C\cite{he2022vrl}). It is noted that VRL still uses a 2D model for frame-by-frame feature extraction and an aggregation layer to get the clip features. However, our 3D-CSL extract clip features through the entire network at once.

\vspace{0.2em}
\noindent\textbf{(1) Retrieval Performance.} As Table\ref{tab:sota} shows, our 3D-CSL achieves the best performance on clip-level methods. Even though we do not use complex editing transformations like VRL, our models trained on base transformation and ShotMix still learn the robustness representation. Compared with frame-level methods on FIVR-200K, our 3D-CSL outperforms all of them except VisiL and VRL-F on DSVR and ISVR tasks. We conjecture that these mAPs are decreased due to the 8 times higher compression rate of the clip-level features, which aggregates 8 frames information into one embedding.


\indent Furthermore, on the most challenging ISVR task, 3D-CSL can achieve \textbf{0.711} mAP, outperforming all prior methods. In this task, some reference videos neither temporally nor spatially overlap with the queries, which greatly increases the retrieval difficulty. The superior result indicates that our model has a powerful ability for long-range spatiotemporal modeling. 

\vspace{0.2em}
    \noindent\textbf{(2) Retrieval Speed and Storage.} Comprehensive performance comparison on retrieval speed and storage is shown in Fig\ref{fig:performance}. All models are tested on Nvidia 2080Ti. Our 3D-CSL achieves the best trade-off between efficiency and effectiveness. The video-level methods require minimal feature storage and retrieval cost, but their low accuracy is insufficient for large-scale accurate retrieval. Compared with these methods, our method achieves a significant improvements by 30.9\%, 28.2\%, and 23.8\% mAPs on FIVR-200K from table\ref{tab:sota}. Compared with the frame-level methods, which require $\mathcal{O}(NM)$ complexities for each pair in Eq\ref{equ:top3cs}, our clip-level method performs only $\mathcal{O}(\displaystyle{NM / 64})$ complexities per pair. Besides, the size of space for storing clip-level features is also decreased by 8 times.

\vspace{-1em}
\subsection{Visualization of Learned Features}
\begin{figure}[h]
\vspace{-0.8em}
  \centering
  \centerline{\includegraphics[width=8.5cm]{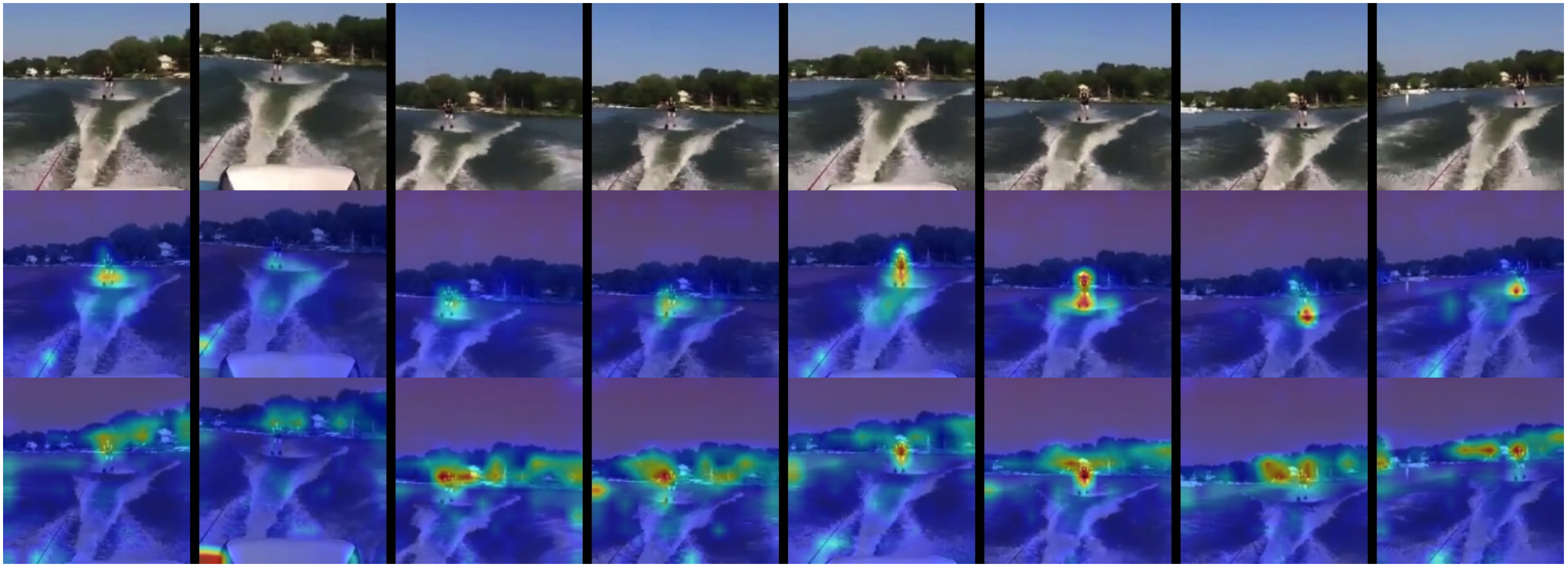}}
\vspace{-1em}
\caption{Visualization result on Kinetics400. (Top) the input clip; (Middle) attention of classification features; (Bottom) attention of similarity features.}
\vspace{-1em}
\label{fig:res}
\end{figure}

\vspace{-0.5em}
Our compact pipeline enables us to explore the inherent learning pattern of similarity learning. For better understanding, we compare it with the supervised action classification task\cite{timesformer}. By visualizing the attention flow of learned features with \cite{abnar2020quantifying-attention-vis}, we observe a significant difference between the two tasks. 
As Fig\ref{fig:res} shows, the classification features focus more on relevant objects for reasoning while ignoring background. In contrast, the similarity features have wider attention spans, which is helpful for the model to distinguish the fine-grained details between videos.



\vspace{-1.2em}
\section{Conclusion}
\vspace{-0.2em}

This paper introduces 3D-CSL, a new pipeline for modeling clip-level features for the NDVR task. It shows potential for long-range video reasoning and speed \& storage advantages. Although the clip-level features are more compact than frame features, they can achieve SOTA results in ISVR. Furthermore, the proposed self-supervised training strategy boosts 3D-CSL for video retrieval. First, the 3D model is pretrained by the proposed PredMAE for learning long-range dependencies more sufficiently. Then, the proposed FCS loss enables the model to distinguish the relative magnitudes of positive similarities, and the video-specific ShotMix strengthens the model's robustness facing multi-shot videos.


\vfill
\pagebreak



\bibliographystyle{IEEEbib}
\begin{spacing}{0.85} 
    \bibliography{refs}
\end{spacing}

\end{document}